\documentstyle[cogsci94,psfig]{article}
\input epsf

\title{ Towards a Principled Representation
of Discourse Plans}

\author{
{\Large\bf R.~Michael Young} \\
Intelligent Systems Program \\
University of Pittsburgh \\
Pittsburgh, PA, 15260 \\
{\tt myoung+@pitt.edu} \\
\And
   {\Large\bf Johanna D. Moore}
   \\Department of Computer Science and
   \\Learning Research and Development Center
   \\University of Pittsburgh
   \\Pittsburgh, PA 15260
   \\{\tt jmoore@cs.pitt.edu}
\And
{\Large\bf Martha E. Pollack}
   \\Department of Computer Science and
   \\Intelligent Systems Program
   \\University of Pittsburgh
   \\Pittsburgh, PA 15260
   \\{\tt pollack@cs.pitt.edu}
}

\newenvironment{DescriptionEx}
	{\begin{list} {}%
		      {\setlength{\leftmargin}{5em}
		       \setlength{\labelwidth}{2em}
		       \setlength{\labelsep}{1em}
	               \setlength{\topsep}{1ex}
                       \setlength{\rightmargin}{5em}
		      }%
	}%
	{\end{list}}

\begin{document}
\bibliographystyle{cogsci}

\maketitle

\begin{abstract}
We argue that discourse plans must capture the intended causal and
decompositional relations between communicative actions.  We present a
planning algorithm, DPOCL, that builds plan structures that properly
capture these relations, and show how these structures are used to
solve the problems that plagued previous discourse planners, and allow
a system to participate effectively and flexibly in an ongoing
dialogue.
\end{abstract}

\section{Introduction}

The close connection between discourse and intention is by now nearly
universally accepted: generating discourse is an intentional activity,
the structure of discourse reflects the structure of the participants'
intentions, and understanding discourse involves, at least in part,
recognizing the intentions of the language producer.  Researchers
working both on generation and interpretation are wont to exhibit
``discourse plans'' that represent the intentions of language users.
However, there has been much confusion about exactly what constitutes
a discourse plan, and what kind of algorithms should process them.
Most of the work in computational linguistics has built on plan
representations and planning algorithms that are at least a decade
old---representations and algorithms that suffer from being
unprincipled and difficult to analyze.  These difficulties have
spilled over into the NL systems that rely on them.  Yet within the
past few years, the literature on AI planning has grown significantly,
and the older representations and algorithms have been reanalyzed and
replaced with cleaner representations and algorithms whose formal
properties are amenable to careful analysis.

In this paper, we illustrate some of the problems that arise from
using these old plan representations and planning algorithms.  We then
show how more recent planning algorithms,  called
partial-order causal link (POCL) planners \cite{SNLP,UCPOP}, can be
used to generate discourse plans.  The particular planning
algorithm we use is DPOCL, an algorithm that introduces action
decomposition into a POCL framework
\cite{DPOCL}.  We show that the discourse plans produced by the
DPOCL algorithm properly capture both the intended causal and
decompositional relations among the communicative actions, and
thereby solve the problems of earlier systems and allow a
language-processing system to participate effectively and flexibly in
an ongoing dialogue.

\section{Previous Approaches}

Discourse is typically viewed as having a hierarchical structure and
therefore many discourse planners are based on the original
NOAH \cite{NOAH} model of hierarchical planning
\cite{AppeltBook,CawseyBook,HovyNLGW88Book,MayburyIJMMS,MooreParisCL}.
These systems rely on customized planning algorithms with procedural
semantics for the purposes of solving specific text-planning problems.
The informal construction of these systems and their application to
particular problems have resulted in successful text generation for
limited domains and text types, while obscuring the undesirable
properties of the algorithms.  However, careful analysis of these
programs shows that there is nothing in their semantics to prevent
them from generating incorrect plans, generating plans with redundant
steps, or failing to find plans in situations where they exist.
To the extent that these planners have been able to avoid these problems,
they have done so by severely limiting the expressive power of action
descriptions and/or requiring the designer of action descriptions to
handcraft each description to fit correctly into the {\em ad hoc} semantics
of the specific plan for which the action is intended.

Within the planning literature, it has been noted that there are two
different ways in which component actions of a plan may be related: an
action ACT1 may provide causal support for another action ACT2 (i.e.,
ACT1 establishes a precondition of ACT2) or an action ACT1 may be part
of the decomposition of ACT2.  Similar distinctions have been noted in
the NL literature, e.g., Grosz and Sidner's
\shortcite{GroszAndSidnerCL} distinction between
satisfaction-precedence and dominance and Pollack's
\shortcite{PollackIIC} distinction between enablement and generation.

The main problem with most previous discourse planning systems is that
they have not adequately represented both the causal and
decompositional relations between actions in a discourse plan.  That
is, they do not reason about interactions between the effects of
actions in the plan.  More specifically, they do not reason about the
establishment of preconditions, or the possibility that one step in
the plan may accidentally undo or obviate the effect of another step.
Moreover, in cases where they perform decomposition, they do not
reason about the relationship between the effects of actions in a
subplan and the effects of their parent action.

\begin{figure}
\rule{3.4in}{0.3mm}
\centerline{\psfig{figure=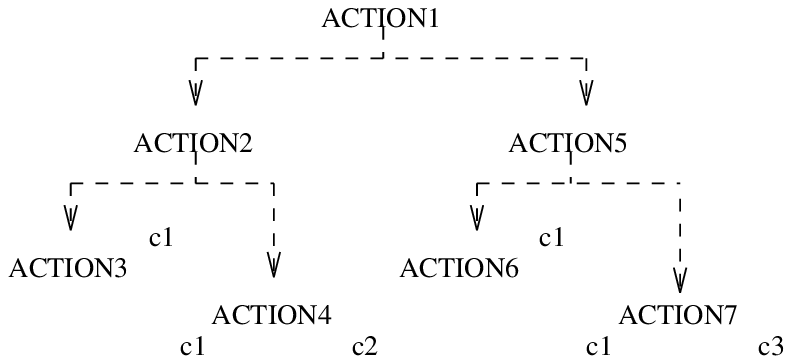}}
\caption{Schematic Discourse Plan Illustrating a Redundant Step}
\label{fig:One}
\rule{3.4in}{0.3mm}
\
\vspace{2ex}
\centerline{\psfig{figure=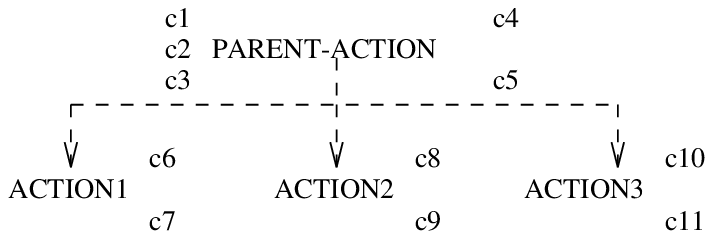}}
\caption{Schematic Discourse Plan Illustrating Parent/Subplan Effects}
\label{fig:Two}
\rule{3.4in}{0.3mm}
\end{figure}

To illustrate two of these problems, consider the discourse plans
shown schematically in Figures~\ref{fig:One} and \ref{fig:Two}.  In
these figures, conditions, denoted by the $c_i$, appearing to the left
of an action denote its preconditions and those appearing to the right
of an action denote its effects.  These plans have structure that is
typical of those produced by most previous discourse planning systems
\cite{CawseyBook,HovyNLGW88Book,MayburyIJMMS,MooreParisCL}.
Figure~\ref{fig:One} shows a plan where the effect $c1$ is established
by two different actions occurring in different subtrees of the plan.
This can occur because these planners do not consider the roles that
previous actions' effects can play in satisfying the preconditions of
subsequent discourse actions.\footnote{Appelt \shortcite{AppeltBook}
would solve this problem with critics, i.e., {\em ad hoc} procedures that
check for certain types of plan interactions.} Thus, they cannot
detect when an action added to establish one particular condition may
serendipitously satisfy conditions of other steps in the plan.  This
failure may lead to the generation of texts that are (unintentionally)
redundant or repetitive.  An analogous, and possibly even more
damaging, problem may result when these systems fail to notice that
one action undoes the effect of another.

Figure~\ref{fig:Two} shows a plan where there is no explicit
connection between the effects established by the parent action ($c4$
and $c5$) and those established by its subplan ($c6$ through $c11$).
Previous approaches only represent the relationship between actions at
different levels; they fail to capture the relationship between the
effects of those actions.  In Figure~\ref{fig:Two}, the top-level goal
is $c4 \wedge c5$.  Suppose that $c6$ unifies with $c4$, and that
$c8$, $c9$, and $c10$ together have a consequence that unifies with
$c5$.  In this case, $c7$ and $c11$ are side effects of choosing the
decomposition of the PARENT-ACTION into ACTION1, ACTION2 and ACTION3.
This fact, however, is not captured in the discourse plan of
Figure~\ref{fig:Two}.  Hence a system relying on this plan could not
distinguish intended effects from side effects, and so would be unable
to determine that the failure of $c6$ warrants a different response
than the failure of $c7$.

In short, these systems cannot, in general, determine how discourse
actions are related to one another.  Yet, as we will illustrate in
the next section, understanding the intended relations between
discourse actions is crucial to effective language generation.

\section{The Significance of Discourse Plans}

Consider the following sample
discourse, a fragment of a political discussion between two
participants, Sharon (S) and Harry (H).

\begin{DescriptionEx}
\item[S:] Wiggins will vote no on NAFTA.  She's an ally of
the unions.  Her district is heavily industrial.
\end{DescriptionEx}

A plausible and typical analysis of this discourse is that Sharon's
primary intention is to convince Harry that Wiggins will vote no on
NAFTA.  To achieve this goal, Sharon asserts the proposition in
question (that Wiggins will vote no on NAFTA) and then supports it by
claiming that Wiggins is an ally of the unions.  To convince Harry of
this later claim, Sharon supports it by claiming that Wiggins's
district is heavily industrial.

Now consider these possible alternative responses by Harry to Sharon's
statement:

\begin{DescriptionEx}
\item[H1:]  I didn't think her district was industrial.

\item[H2:]  Lots of representatives from industrial districts vote against the
union.

\item[H3:]  Well, she's certainly pro-union, but I didn't think her district
was industrial.

\item[H4:]  Well, she's certainly pro-union, but lots of representatives from
industrial districts support NAFTA.

\item[H5:]  I didn't think her district was industrial.  And besides,
lots of representatives from industrial districts support
NAFTA.
\end{DescriptionEx}

\label{sec:replies}

How is Sharon to determine an appropriate response to these replies?
As we have pointed out \cite{MooreParisCL,MoorePollackCL},
Sharon's response must take account of what Harry's reply reveals
about which parts of Sharon's discourse plan were successful.


For example, in H1 Harry's failure to believe that Wiggins's district
is industrial blocks the support that this claim would have provided
to convince Harry of Wiggins's pro-union position. At this point
Sharon has several options.  She may try to convince Harry that
Wiggins's district is, in fact, industrial.  Alternatively, she may
find some other support for the claim that Wiggins is pro-union or she
may find some other means to support the anti-NAFTA claim altogether.


Implicit in Sharon's initial statement was her belief that Harry believed
that, as a rule, a representative's position on labor is determined by
the industrial make-up of her district.  Together with Sharon's claim
that Wiggins's district is industrial, this rule would have provided
support to convince Harry of Wiggins's pro-union position.  H2
indicates that the support for Sharon's claim that Wiggins is
pro-union has failed.  Sharon must either find an alternate
discourse strategy for supporting it or must find some other means to
support the anti-NAFTA claim.  Notice the difference between H1 and
H2.  A plausible response to H1, but not H2, is to reestablish the
proposition that Wiggins's district is industrial.


In H3 as in H1, Harry expresses doubt that Wiggins comes
from an industrial district. However, he also indicates that he
believes that Wiggins is pro-union.  Sharon's intention to get Harry
to believe that Wiggins's district is industrial was not achieved.
Consequently we may infer that her intention to get him to believe
that Wiggins is pro-union also failed.  However, Sharon need not try
to provide alternate support for either her pro-union or anti-NAFTA
claims.  This is because Sharon's intention to convince Harry that
Wiggins's district is industrial was held in service of the intention
to get Harry to believe in Wiggins's pro-union position.  That is,
there was a causal connection between the industrial-district
intention and the pro-union intention; because Harry's response
explicitly indicates that the pro-union intention was achieved, the
outcome of those intentions which served as preconditions to it or as
effects in subplans of it can be ignored.\footnote{If, on the other
hand, it is important to Sharon (for some other reason) that Harry also
believe that Wiggins comes from an industrial district, then she may
need to reconvince him of this.}


Responses H4 and H5 are variations of H1 through H3.  Their analysis
is left to the reader.

As can be seen in these examples, a wide range of responses to Harry's
replies are possible.  Each of Harry's replies provides feedback about the
outcome of a small subset of Sharon's intentions.  In order to respond
appropriately, Sharon must be able to determine what implications this
feedback has on the ultimate success of her other intentions.

\section{A Discourse Plan for Our Example}

\begin{figure*}
\rule{7.0in}{0.3mm}
\centerline{\psfig{figure=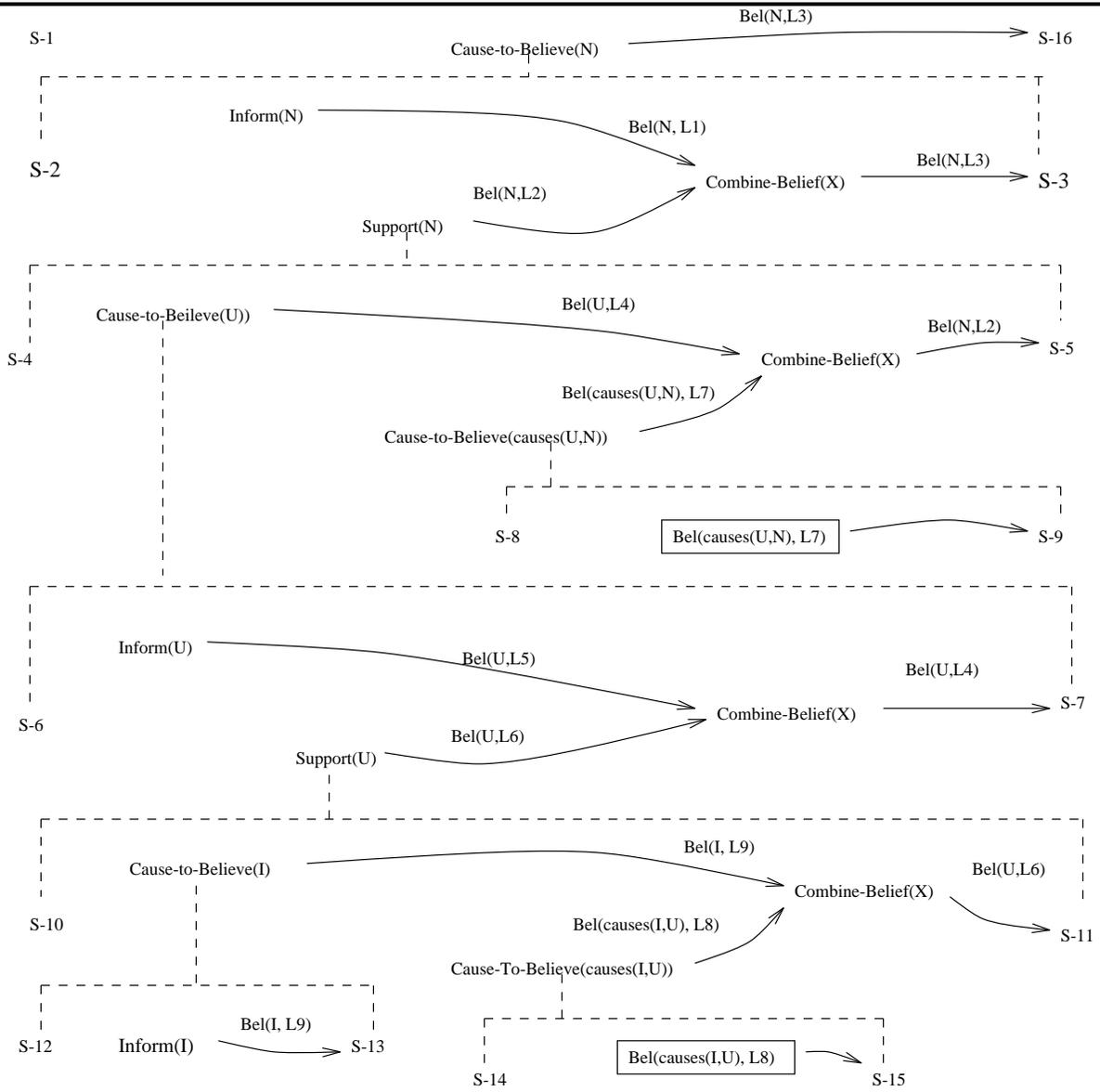}}
\caption{An Example Discourse Plan}
\label{fig:Three}
\rule{7.0in}{0.3mm}
\end{figure*}

We now describe how the DPOCL system represents Sharon's utterance under the
analysis given above; see  Figure~\ref{fig:Three}.

The manner in which a hearer combines the information in an utterance
with his prior beliefs is critical to the generation of the utterance.
Most previous work has made use of highly simple models of this
process: for instance, it has assumed that the effect of asserting a
proposition $p$ is either that the hearer believes or does not believe
$p$. In fact, a speaker may go to great lengths to convince the hearer
of the truth of a proposition.  She may first assert it, then support
it, and then provide support for the intermediate statement.  In such
a case, the speaker presumably believes that the combination of
utterances is what leads the hearer to accept the main proposition.  A
complete model of this phenomenon is beyond the scope of this paper;
we hint at it by representing the combination of multiple partial
beliefs with the action Combine-Belief$(\vec{x})$, where $\vec{x}$ is
a vector of relevant beliefs.  The strength of belief $L$ that a
hearer has in a particular proposition $P$ is represented informally
by the formula Bel$(P,L)$.

In Figure~\ref{fig:Three} we abbreviate propositions as follows: $N$
represents Wiggins will vote No on NAFTA, $U$ represents Wiggins is
pro-union and $I$ represents Wiggins comes from an industrial
district.  Those conditions surrounded by boxes are true in the
initial state -- causal link arcs connecting them to the initial state
are omitted for clarity.

The DPOCL data structure for representing  plans consists
of five components:

\begin{itemize}

\item {\bf Steps:}  Each discourse action in the plan is represented by
a step.  These steps are the nodes in the plan graph. Steps are
instantiated from action operators representing the action's
preconditions and effects. Steps may be composite, representing
abstract actions like Cause-to-Believe$(U)$, or primitive, representing
those actions that are directly executable by the system, such as
Inform$(I)$.

\item {\bf Decomposition Links:}  Decomposition links connect a parent step to
the initial and final steps of the subplan that achieves the parent
step's effects.  The decomposition links are shown using dashed arcs;
they capture the hierarchical structure of the plan.

\item {\bf Causal Links:}  Causal links connect two steps when the first step
establishes a precondition for the second step.  They are shown using
solid arcs and are labeled with the effects that they contribute.

\item {\bf Ordering Constraints:} The set of ordering constraints defines a
partial temporal ordering over the steps in the plan.  For
readability, these constraints are not shown in
Figure~\ref{fig:Three}.

\item {\bf Binding Constraints:}  The set of binding constraints provide
codesignation relationships for variables occurring in the steps of
the plan.  For readability, all variables in the plan shown in
Figure~\ref{fig:Three} have been replaced by object constants specified
by the plan's binding constraints.

\end{itemize}

DPOCL uses the standard technique of encoding the initial conditions
and the goals of a planning problem as the effects of a null initial
action and the preconditions of a null final action, respectively.
Similarly, in an action decomposition, there is a null initial action
that has as its effects exactly the preconditions of its parent
action, and a null final action that has as its preconditions the
effects of its parent.
The DPOCL planner attempts to achieve the preconditions of a subplan's
final step in the same manner as it achieves all other unsatisfied
preconditions.  In this way we guarantee that the effects of every
composite action are achieved by the steps in its subplan.
Furthermore, the exact relationship between the actions in a subplan
and the establishment of those effects is made explicit.

By analyzing the causal and decompositional structure of the discourse
plan, we can determine an appropriate response for each of Harry's
replies discussed above:

\label{sec:plan-example}

\begin{itemize}


\item {\bf H1:} In this case, the effect Bel$(I,L9)$ asserted by the
Inform$(I)$ was not achieved.  From our representation, it is possible
to trace a path of failed effects from Inform$(I)$ across causal links
and up decompositional links to Cause-To-Believe$(I)$, Support$(U)$
and eventually to Cause-To-Believe$(N)$.  Using this information, the
system can determine that appropriate responses to H1 can be generated
by trying to convince Harry that Wiggins's district is, in fact,
industrial (i.e, replanning the subtree rooted at the node
Cause-To-Believe$(I)$, most likely by providing support for $I$),
finding some other support for the claim that Wiggins is pro-union
(i.e, replanning the subtree rooted at the node Support$(U)$), or
finding some other means to support the anti-NAFTA claim altogether
(i.e, replanning the subtree rooted at the node Support$(N)$).


\item {\bf H2:}  The DPOCL plan in Figure~\ref{fig:Three} is predicated
on the truth of Bel(causes$(I,U), L8)$, i.e., that this proposition is
an effect of the initial step.  In H2, Harry reveals that this
proposition is false.  As in the previous case, an appropriate
response results from the re-planning of subtrees whose execution is
affected by this failure. Specifically, those subtrees rooted at
Cause-To-Believe(causes$(I,U))$ across causal links and up
decompositional links to Support$(U)$ and eventually to
Cause-To-Believe$(N)$.  Note that this does not include the subtree
rooted at Cause-to-Believe$(I)$ and thus, unlike in H1, the system
will not attempt to reestablish the proposition that Wiggins's
district is industrial.


\item {\bf H3:} In this case, the speaker is given more information
about the success of the original plan.  As in H1, the effect
Bel$(I,L9)$ is not achieved.  However, here Harry also indicates that
the effect Bel$(U,L4)$ of the step Cause-To-Believe$(U)$ {\em has\/}
been achieved.  Cause-To-Believe$(U)$ lies along the only causal path
from Inform$(I)$ to the plan's final step.  Since it achieved its
intended effect, re-planning any of its subplans is unnecessary.

\end{itemize}

Although this example did not explicitly illustrate
how our representation addresses cases where action descriptions have
multiple effects, it is clear our model can handle such cases
appropriately.  Our solution rests on the fact that our model makes a
clear distinction between effects of discourse actions that play a
role in achieving the top-level goals of the discourse plan and
effects that are not important for achieving the agent's ultimate
goals (i.e., side effects).

\section{How DPOCL Creates Discourse Plans}

So far we have focused on the representation used by DPOCL.  We now
briefly describe how the DPOCL algorithm works.  In DPOCL, the process
of creating a completed plan involves iterating through a loop that
chooses between refining the current plan decompositionally (expanding
a composite action by adding its subactions to the plan) or refining
the plan causally (choosing some action's unsatisfied precondition and
adding a new action and the causal link establishing it).
Figure~\ref{fig:Four} summarizes the DPOCL planning algorithm.  For
more details of the algorithm and a discussion of its formal
properties, see \cite{DPOCL}.

The representation of each action is separated into two parts
corresponding to the causal and decompositional roles the action
plays: the action operator, and a possibly empty set of decomposition
operators.  The action operator captures the action's preconditions
and effects.  These preconditions and effects are sets of first-order
quantified sentences similar to the typical precondition and
add/delete lists of STRIPS \cite{STRIPS}.  Each decomposition operator
represents a single-layer expansion of a composite step, essentially
providing a partial specification for the subplan that achieves the
parent step's effects given its preconditions. In addition to
specifying the steps in the subplan, the decomposition operator
specifies any variable binding and temporal ordering constraints
between the steps, and the causal links between steps of the subplan
that enable them to establish the parent step's effects.

\begin{figure}[t]
\rule{3.4in}{0.3mm}

\vspace{.15in}

{\footnotesize

{\bf  Termination:} If the plan is inconsistent, then backtrack.
Otherwise, remove unused step and return the plan.

\vspace{2ex}

{\bf  Plan Refinement:} Non-deterministically do one of the following:

\begin{enumerate}

\item {\bf  Causal Planning:}

\begin{enumerate}

\item {\bf  Goal Selection:} Nondeterministically select a goal.

\item {\bf  Operator Selection:} Add a step to the plan that adds an effect
that can be unified with the goal (either by instantiating the step
from the operator library or by finding a step already in the plan).
If no such step exists, backtrack.  Otherwise, add the binding constraints
required for the conditions to unify, an  ordering constraint that orders
the new step before the goal step and add the causal link between the two.

\end{enumerate}

\item {\bf  Decompositional Planning:}

\begin{enumerate}

\item {\bf  Action Selection:} Nondeterministically select some
unexpanded composite step in the plan.

\item {\bf  Decomposition Selection:} Nondeterministically chose an
appropriate decomposition schema for this action whose constraints are
satisfied.  Add the steps and
subplan components of the decomposition schema to the plan and update the list
of decomposition links to indicate the new subplan.

\end{enumerate}

\end{enumerate}

\noindent{\bf  Threat Resolution:}  Find any step that might threaten to undo
any causal link.  For every such step, nondeterministically do one of
the following:

\begin{itemize}

\item {\bf  Promotion} If possible, move the threatened steps to occur before
the threat in the plan.

\item {\bf  Demotion} If possible, move the threatened steps to occur after
the threat in the plan.

\item {\bf  Separation} If possible, add binding constraints on the
steps involved so that no conflict can arise.

\end{itemize}

\noindent{\bf  Recursive Invocation}  Call the planner recursively with
the new plan structure.

}   
\caption{The DPOCL Algorithm}
\label{fig:Four}
\rule{3.4in}{0.3mm}
\end{figure}


The formal specification of the DPOCL algorithm relies on
nondeterministic choice to guide its search through the space of
partial plans.  Each choice is recorded, and backtracking occurs when
appropriate.  Nondeterministic choice is specified in order to allow
DPOCL implementations to specify domain-dependent search control.  As
long as search control heuristics guarantee that all possible choices
will be explored, the implementation remains complete.

As a result of adding steps to a plan, newly created steps may
introduce {\em threats} to existing causal links.  A step $A$ threatens a
causal link between two steps $B$ and $C$ when $A$ might occur between
$B$ and $C$ and one of $A$'s effects might undo the condition
established in the causal link.  To ensure that no causal links are
undone, each threat is dealt with before planning proceeds, either
by ordering the steps so that the threatening step cannot occur between
the two causally-linked steps or by restricting the variable bindings
of the steps to eliminate harmful interactions.  This process is
iterative, since each modification to resolve a threat may introduce
new ones.

In the example discussed earlier, the DPOCL planner is
invoked with the partial plan consisting of the null initial action
$S$-$1$, whose effects are Bel(causes$(U,N), L7)$ and
Bel(causes$(I,U), L8)$, and the null final action $S$-$16$, whose
only precondition is Bel$(N,L3)$.  By iterating through the DPOCL loop
shown in Figure~\ref{fig:Four}, the plan shown in
Figure~\ref{fig:Three} is constructed.  This plan is completed, that
is, the preconditions of all actions have been established by causal
links, there are no threats to any of these links, and all composite
actions have been decomposed into subplans terminating in executable
actions at the leaf nodes.

This plan makes explicit the causal connections between each effect
and the precondition that relies upon it.  Similarly, the
decomposition links make explicit the manner in which actions in a
subplan establish the effects of the parent step.  This representation
makes it possible for a system playing the role of Sharon to respond
appropriately to each of Harry's replies as described earlier.


\section{Conclusions}

In this paper, we have presented a structure for discourse plans that
draws on state-of-the-art AI planning research.  Both the plan
representation and the discourse planning algorithm that we use to
construct it have a well-defined semantics whose formal
properties can be analyzed \cite{DPOCL}.  Further, we have shown
how DPOCL discourse plan structures can be used for determining
appropriate responses to utterances that indicate a failure of some
part of the discourse plan.

\section{Acknowledgements}

The authors would like to thank the anonymous reviewers for their
helpful comments.

The research described in this paper was supported by the Office of
Naval Research Cognitive and Neural Sciences Division (Grant Number:
N00014-91-J-1694) and by the National Science Foundation and Advanced
Research Projects Agency under Grant IRI-9304961 (Integrated
Techniques for Generation and Interpretation).  Young is supported by
a grant from ONR under the FY93 Augmentation of Awards for Science and
Engineering Research Training (ASSERT) Program.  Pollack is supported
by the Air Force Office of Scientific Research (Contract
F49620-92-J-0422), by the Rome Laboratory (RL) of the Air Force
Material Command and the Defense Advanced Research Projects Agency
(Contract F30602-93-C-0038), and by an NSF Young Investigator's Award
(IRI-9258392).



\end{document}